\documentclass[sigconf,letter]{acmart}

\usepackage{array}
\usepackage{booktabs} 
\usepackage{algorithm}
\usepackage{xspace}
\usepackage{relsize}
\usepackage[noend]{algorithmic}
\usepackage{multirow}
\setcopyright{rightsretained}
\usepackage[T1]{fontenc}
\newcommand{\DEDDQN}{\mbox{DE-DDQN}\xspace}
\newcommand{\FEmax}{\ensuremath{FE^\text{max}}\xspace}

\newcommand{\MANUEL}[1]{}
\renewcommand{\MANUEL}[1]{\protect\footnote{\textcolor{red}{\noindent\textbf{[MANUEL: #1]}}}}

\newcommand{\DIMITAR}[1]{}
\renewcommand{\DIMITAR}[1]{\protect\footnote{\textcolor{blue}{\noindent\textbf{[DIMITAR: #1]}}}}

\newcommand{\MUDITA}[1]{}
\renewcommand{\MUDITA}[1]{\protect\footnote{\textcolor{green}{\noindent\textbf{[MUDITA: #1]}}}}
\newcommand{\hide}[1]{}

\DeclareMathOperator*{\argmax}{arg\,max}

\copyrightyear{2019} 
\acmYear{2019} 
\setcopyright{acmcopyright}
\acmConference[GECCO '19]{Genetic and Evolutionary Computation Conference}{July 13--17, 2019}{Prague, Czech Republic}
\acmBooktitle{Genetic and Evolutionary Computation Conference (GECCO '19), July 13--17, 2019, Prague, Czech Republic}
\acmPrice{15.00}
\acmDOI{10.1145/3321707.3321813}
\acmISBN{978-1-4503-6111-8/19/07}

\begin{document}
\title[Deep Reinforcement Learning Based Parameter Control in Differential Evolution]{Deep Reinforcement Learning Based Parameter Control\\ in Differential Evolution}


\author{Mudita Sharma}
\affiliation{%
  \institution{University of York}
  \city{York} 
  \state{U.K.}}
\email{ms1938@york.ac.uk}

\author{Alexandros Komninos}
\affiliation{%
  \institution{University of York}
  \city{York} 
  \state{U.K.}} 
\email{alexandros.komninos@york.ac.uk}

\author{Manuel L\'opez-Ib\'a\~nez}
\affiliation{%
  \institution{University of Manchester}
  \city{Manchester} 
  \country{U.K.}}
\email{manuel.lopez-ibanez@manchester.ac.uk}

\author{Dimitar Kazakov}
\affiliation{%
  \institution{University of York}
  \city{York} 
  \country{U.K.}}
\email{dimitar.kazakov@york.ac.uk}

\renewcommand{\shortauthors}{M. Sharma et al.}

\begin{abstract}
Adaptive Operator Selection (AOS) is an approach that controls discrete parameters of an Evolutionary Algorithm (EA) during the run. In this paper, we propose an AOS method based on Double Deep Q-Learning (DDQN), a Deep Reinforcement Learning method, to control the mutation strategies of Differential Evolution (DE). The application of DDQN to DE requires two phases. First, a neural network is trained offline by collecting data about the DE state and the benefit (reward) of applying each mutation strategy during multiple runs of DE tackling benchmark functions. We define the DE state as the combination of 99 different features and we analyze three alternative reward functions. Second, when DDQN is applied as a parameter controller within DE to a different test set of benchmark functions, DDQN uses the trained neural network to predict which mutation strategy should be applied to each parent at each generation according to the DE state. Benchmark functions for training and testing are taken from the CEC2005 benchmark with dimensions 10 and 30. 
We compare the results of the proposed \DEDDQN algorithm to several baseline DE algorithms using no online selection, random selection and other AOS methods, and also to the two winners of the CEC2005 competition. The results show that \DEDDQN outperforms the non-adaptive methods for all functions in the test set; while its results are comparable with the last two algorithms.
  
\end{abstract}

%
%



\maketitle

\newcommand{\maxgen}{\ensuremath{\textit{gen}}\xspace}
\newcommand{\xbsf}{\ensuremath{\vec{x}_\text{bsf}}\xspace}
\newcommand{\xbest}{\ensuremath{\vec{x}_\text{best}}\xspace}
\newcommand{\fbsf}{\ensuremath{f_\text{bsf}}\xspace}
\newcommand{\fwsf}{\ensuremath{f_\text{wsf}}\xspace}
\newcommand{\stdev}{\ensuremath{\textrm{std}}}
\newcommand{\maxstdev}{\ensuremath{\textrm{std}^\text{max}}}
\newcommand{\fdim}{\ensuremath{\textit{dim}_f}}
\newcommand{\maxfdim}{\ensuremath{\textit{dim}^\text{max}}}
\newcommand{\stagcount}{\ensuremath{\textit{stagcount}}\xspace}
\newcommand{\maxdist}{\ensuremath{\textit{dist}^\text{max}}\xspace}
\newcommand{\distance}[1]{\ensuremath{\textit{dist}(#1)}\xspace}
\newcommand{\Nsucc}{\ensuremath{N^\text{succ}}}
\newcommand{\Nfail}{\ensuremath{N^\text{fail}}}
\newcommand{\Ntot}{\ensuremath{N^\text{tot}}}
\newcommand{\SuccRate}{\textit{SR}}
\newcommand{\op}{\textit{op}}
\newcommand{\OM}{\textit{OM}}
\newcommand{\BestOM}{\textit{OM}^\text{best}}
\newcommand{\xii}{\ensuremath{\vec{x}_i}\xspace}
\newcommand{\xj}{\ensuremath{\vec{x}_j}\xspace}
\newcommand{\ui}{\ensuremath{\vec{u}_i}\xspace}
\newcommand{\rtarget}{\ensuremath{r^\text{target}}\xspace}

\section{Introduction}

Evolutionary algorithms for numerical optimization come in many variants
involving different operators, such as mutation strategies and types of
crossover. In the case of differential evolution (DE)~\citep{StoPri1997:de},
experimental analysis has shown that different mutation strategies perform
better for specific optimization problems~\citep{MezVelCoe2006} and that
choosing the right mutation strategy at specific stages of an optimization
process can further improve the performance of DE~\cite{FiaRosScho2010comp}. As
a result, there has been great interest in methods for controlling or selecting
the value of discrete parameters while solving a problem, also called
\emph{adaptive operator selection (AOS)}.

In the context of DE, there is a finite number of mutation strategies
(operators) that can be applied at each generation to produce new solutions
from existing (parent) solutions. An AOS method will decide, at each
generation, which operator should be applied, measure the effect of this
application and adapt future choices according to some reward function. An
inherent difficulty is that we do not know which operator is the most useful at
each generation to solve a previously unseen problem. Moreover, different
operators may be useful at different stages of an algorithm's run.

There are multiple AOS methods proposed in the
literature~\citep{KarHooEib2015:tec,AleMos2016slr,GonFiaCai2010adaptive} and
several of them are based on \emph{reinforcement learning (RL)} techniques such
as probability matching~\citep{FiaSchoSeb2010toward,ShaLopKaz2018ppsn},
multi-arm bandits~\citep{GonFiaCai2010adaptive}, $Q(\lambda)$
learning~\citep{PetEve2002control} and
SARSA~\cite{CheGaoChen2005scga,EibHorKow2006rl,SakTakKaw2010}, among
others~\cite{KarEibHoo2014generic}. These RL methods use one or few features to
capture the state of the algorithm at each generation, select an operator to be
applied and calculate a reward from this application. Typical state features
are fitness standard deviation, fitness improvement from parent to offspring,
best fitness, and mean fitness~\cite{EibHorKow2006rl,KarEibHoo2014generic}.
Typical reward functions measure improvement achieved over the previous
generation~\cite{KarEibHoo2014generic}. Other parameter control methods use an
offline training phase to collect more data about the algorithm than what is
available within a single run. For example, \citet{KeeAirCyr2001adaptive} uses two
types of learning: table-based and rule-based. The learning is performed during
an offline training phase that is followed by an online execution phase where
the learned tables or rules are used for choosing parameter values. More
recently, \citet{KarSmiEib2012generic} trains offline a feed-forward neural
network with no hidden layers to control the numerical parameter values of an
evolution strategy. To the best of our knowledge, none of the AOS methods that
use offline training are based on reinforcement learning.

In this paper, we adapt \emph{Double Deep Q-Network
  (DDQN)}~\cite{HassGueSil2016deep}, a deep reinforcement learning technique
that uses a deep neural network as a prediction model, as an AOS method for DE.
The main differences between DDQN and other RL methods are the possibility of training
DDQN offline on large amounts of data and of using a larger number of features to
define the current state. When applied as an AOS method within DE, we first run
the proposed \DEDDQN algorithm many times on training benchmark problems by
collecting data on $99$ features, such as the relative fitness of the
current generation, mean and standard deviation of the population fitness,
dimension of the problem, number of function evaluations, stagnation, distance
among solutions in decision space, etc. After this training phase, the \DEDDQN
algorithm can be applied to unseen problems. It will observe the run time value
of these features and predict which mutation strategy should be used at each
generation. \DEDDQN also requires the choice of a suitable reward definition to facilitate learning of a prediction model.
Some RL-based AOS methods calculate rewards per
individual~\cite{PetEve2002control,CheGaoChen2005scga}, while others 
calculate it per generation~\cite{SakTakKaw2010}.
Moreover, reward functions can be designed in different ways depending on the problem at hand. For example, \citet{KarHooEib2015eval} defines and compares four per-generation reward definitions for RL-based AOS methods.
Here, we also find that the reward definition has a strong effect on the
performance of \DEDDQN and, hence, we analyze three alternative reward
definitions that assign reward for each application of a mutation strategy.

As an experimental benchmark, we use functions from the \textsc{cec}2005
special session on real-parameter optimization~\citep{SugHanLia2005cec}. In
particular, the proposed \DEDDQN method is first trained on 16 functions for
both dimensions 10 and 30, i.e., a total of 32 training functions. Then, we run the
trained \DEDDQN on a different set of 5 functions, also for dimensions 10 and
30, i.e., a total of 10 test functions. We also run on these 10 test functions the
following algorithms for comparison: four DE variants, each using a single
specific mutation strategy, DE with a random selection among mutation
strategies at each generation, DE using various AOS methods
(PM-AdapSS~\cite{FiaSchoSeb2010toward}, F-AUC~\cite{GonFiaCai2010adaptive}, and
RecPM-AOS~\cite{ShaLopKaz2018ppsn}), and the two winners of
CEC2005~\cite{SugHanLia2005cec} competition, which are both variants of CMAES:
LR-CMAES (LR)~\cite{AugHan2005lrcmaes} and IPOP-CMAES
(IPOP)~\cite{AugHan2005cec}.

Our experimental results show that the DE variants using AOS completely
outperform the DE variants using a fixed mutation strategy or a random
selection. Although a non-parametric post-hoc test does not find that the
differences between the CMAES algorithms and the AOS-enabled DE algorithms
(including \DEDDQN) are statistically significant, \DEDDQN is the second best
approach, behind IPOP-CMAES, in terms of mean rank.

The paper is structured as follows. First, we give a brief introduction to DE, mutation strategies and deep reinforcement learning. In Sect.~\ref{sec:ddqn}, we introduce our proposed \DEDDQN algorithm, and explain its training and online (deployment) phases. Section~\ref{proposed} introduces the state features and reward functions used in the experiments, which are described in Sect.~\ref{sec:exp}. We summarise our conclusions in Sect.~\ref{sec:conclusion}.

\section{Background}

\subsection{Differential Evolution}\label{lab:DE}
Differential Evolution (DE)~\cite{PriStoLam2005:book} is a population-based
algorithm that uses a mutation strategy to create an offspring solution
$\vec{u}$. A mutation strategy is a linear combination of three or more parent
solutions $\xii$, where $i$ is the index of a solution in the current
population. Some mutation strategies are good at exploration and others at exploitation, and it is well-known that no single strategy performs best for all problems and for all stages of a single run. In this paper, we consider these frequently used mutation strategies:
%
\vspace{-1em}
\begin{center}
\begin{tabular}{@{}r@{\hspace{1ex}}l@{}}
  \text{``rand/1''}:& $\ui = \vec{x}_{r_1} + F \cdot (\vec{x}_{r_2}-\vec{x}_{r_3})$\\
  \text{``rand/2''}:& $\ui = \vec{x}_{r_1} + F\cdot (\vec{x}_{r_2}-\vec{x}_{r_3} + \vec{x}_{r_4}-\vec{x}_{r_5})$\\
  \text{``rand-to-best/2''}:&  $\ui = \vec{x}_{r_1} + F\cdot (\xbest-\vec{x}_{r_1} + \vec{x}_{r_2}-\vec{x}_{r_3} + \vec{x}_{r_4}-\vec{x}_{r_5})$\\
  \text{``curr-to-rand/1''}:& $\ui = \xii + F\cdot (\vec{x}_{r_1}-\xii+\vec{x}_{r_2}-\vec{x}_{r_3})$\\
\end{tabular}
\end{center}
where $F$ is a scaling factor, $\ui$ and $\xii$ are the $i$-{th}
offspring and parent solution vectors in the population, respectively,
$\xbest$ is the best parent in the population, and $r_1$, $r_2$,
$r_3$, $r_4$, and $r_5$ are randomly generated indexes within $[1, NP]$, where
$NP$ is the population size. An additional numerical parameter, the crossover
rate ($CR \in [0,1]$), determines whether the mutation strategy is applied to
each dimension of $\xii$ to generate $\ui$. At least one dimension
of each $\xii$ vector is mutated.

\subsection{Deep Reinforcement Learning}
%
In RL~\cite{SutBar1998reinf}, an agent takes actions in an environment that
returns the reward and the next state. The goal is to maximize the cumulative
reward at each step. RL estimates the value of an action given a state called
\emph{Q-value} to learn a \emph{policy} that returns an action given a state. A
variety of different techniques are used in RL to learn this policy and some of
them are applicable only when the set of actions is
finite. 


When the features that define a state are continuous or the set of states is very large, the policy becomes a function that implicitly maps between state features and actions, as opposed to keeping an explicit map in the form of a lookup table. In \emph{deep reinforcement learning}, this function is approximated by a deep neural network and the weights of the network are optimized to maximize the cumulative reward.

Deep Q-network (DQN)~\cite{MniKavSil2015human} is a deep RL technique that extends
Q-learning to continuous features by approximating a non-linear Q-value
function of the state features using a neural network (NN). The classical
DQN algorithm sometimes overestimates the Q-values of the actions, which leads
to poor policies. Double DQN (DDQN)~\cite{HassGueSil2016deep} was proposed as a way to
overcome this limitation and enhance the stability of the Q-values. DDQN
employs two neural networks: a primary network selects an action and a target
network generates a target Q-value for that action. The target-Q values are
used to compute the loss function for every action during training. The weights
of the target network are fixed, and only periodically or slowly updated to the
primary Q-networks values.

In this work, we integrate DDQN into DE as an AOS method that selects a
mutation strategy at each generation.



\section{\DEDDQN}\label{sec:ddqn}

When integrated with DE as an AOS method, DDQN is adapted as follows. The
environment of DDQN becomes the DE algorithm performing an optimization run for
a maximum of \FEmax function evaluations.  A \emph{state} $s_t$ is a collection
of features that measure static or run time features of the problem being solved
or of DE at step $t$ (function evaluation or generation counter). The actions
that DDQN may take are the set of mutation strategies available
(Sect.~\ref{lab:DE}), and $a_t$ is the strategy selected and applied at step
$t$. Once a mutation strategy is applied, a reward function returns the
estimated benefit (reward) $r_t$ of applying action $a_t$, and the DE run
reaches a new state, $s_{t+1}$. We refer to the tuple $\langle s_t$, $a_t$,
$r_t$, $s_{t+1}\rangle$ as an \emph{observation}.

Our proposed \DEDDQN algorithm operates in two phases. In the first
\emph{training} phase, the two deep neural networks of DDQN are trained on
observations by running the \DEDDQN algorithm multiple times on several
benchmark functions. In a second online (or deployment) phase, the trained DDQN
is used to select which mutation strategy should be applied at each generation
of DE when tackling unseen (or test) problems not considered during the
training phase. We describe these two phases in detail next.


\subsection{Training phase}

In the training phase, DDQN uses two deep neural networks (NNs), namely primary
NN and target NN. The primary NN predicts the Q-values $Q(s_t, a; \theta)$
that are used to select an action $a$ given state $s_t$ at step $t$, while
the target NN estimates the target Q-values $\hat{Q}(s_t, a; \hat{\theta})$
after the action $a$ has been applied, where $\theta$ and $\hat{\theta}$ are the
weights of the primary and target NNs, respectively, $s_t$ is the state vector
of DE, and $a$ is a mutation strategy.

The goal of the training phase is to train the primary NN of DDQN so that it
learns to approximate the target $\hat{Q}$ function. The training data is a
memory of observations that is collected by running \DEDDQN several times on
training benchmark functions. Training the primary NN involves finding its
weights $\theta$ through gradient optimization.

The training process of \DEDDQN is shown in Algorithm ~\ref{algo:ddqn-train}. 
Training starts by running DE with random selection of mutation strategy for a fixed number of steps (\emph{warm-up size}) that generates observations to populate a memory of capacity $N$, which can be different from the warm-up size (line~\ref{line:warmup}). This memory stores a fixed number of $N$ recent observations, old ones are removed as new ones are added.
Once the warm-up phase is over, DE is executed $M$ times, and each run is
stopped after \FEmax function evaluations or the known optimum of the training
problem is reached (line~\ref{line:stop}). For each solution in the population,
the $\epsilon$-greedy policy is used to select mutation strategy, i.e., with
$\epsilon$ probability a random mutation is selected, otherwise the mutation
strategy with maximum Q-value is selected.  Using the current DE state $s_t$,
the primary NN is responsible for generating a Q-value per possible mutation
strategy (line~\ref{line:select}). The use of a $\epsilon$-greedy policy forces
the primary NN to explore mutation strategies that may be currently predicted
less optimal. The selected mutation strategy is applied
(line~\ref{line:mutate}) and a new state $s_{t+1}$ is achieved
(line~\ref{line:newstate}). A reward value $r_t$ is computed by measuring the
performance progress made at this step.


To prevent the primary NN from only learning about the immediate state of this
DE run, 
randomly draw mini batches of observations (line~\ref{line:batches}) from
memory to perform a step of gradient optimization. Training the primary NN with the randomly drawn
observations helps to robustly learn to perform well in the task.

\begin{algorithm}[!t]
  \smaller
  \begin{algorithmic}[1]
  \STATE Initialise parameter values of DE ($F$, $NP$, $CR$) 
  \STATE Run DE with random selection of mutation strategy to initialise memory to capacity
  $N$\label{line:warmup}
  \STATE Initialise Q-value for each action by setting random weights $\theta$ of primary NN
  \STATE Initialise target Q-value $\hat{Q}$ for each action by setting weights $\hat{\theta} = \theta$  of target NN
  \FOR{run $1, \dotsc M$}
  \STATE $t=0$ \\
    \WHILE{$t < \FEmax$  or optimum is reached}\label{line:stop}
     \FOR{$i=1,\dotsc, NP$}
        \IF{rand(0, 1) < $\epsilon$}
            \STATE Randomly select a mutation strategy $a_t$\\
        \ELSE 
        \STATE Select $a_t = \argmax_a Q(s_t, a; \theta)$\label{line:select}
        \ENDIF
        \STATE Generate trial vector $\ui$ for parent $\xii$ using mutation $a_t$\label{line:mutate}
        \STATE Evaluate trial vector and keep the best among $\xii$ and $\ui$\label{line:newstate}
         \STATE Store observation ($s_t$, $a_t$, $r_t$, $s_{t+1}$) in memory
        \STATE Sample random mini batch of observations from memory\label{line:batches}
        \IF{run terminates}
        \STATE $\rtarget = r_t$
        \ELSE
        \STATE $\hat{a}_{t+1} = \argmax_{a} Q(s_{t+1}, a; \theta)$\label{line:pred_a}
        \STATE $\rtarget =  r_t +\gamma \hat{Q}(s_{t+1}, \hat{a}_{t+1}; \hat{\theta})$ \label{line:q-value}
        \ENDIF
        \STATE Perform a gradient descent step on $(\rtarget - Q(s_j, a_j; \theta))^2$ with respect to $\theta$\label{line:train}
        \STATE Every $C$ steps set $\hat{\theta} = \theta$\label{line:sync}
        \STATE $t=t+1$
     \ENDFOR
    \ENDWHILE
\ENDFOR
\RETURN  $\theta$ (weights of primary NN)
\end{algorithmic}
\caption{\DEDDQN training algorithm}\label{algo:ddqn-train}
\end{algorithm}

The primary NN is used to predict the next mutation strategy $\hat{a}_{t+1}$
(line~\ref{line:pred_a}) and its reward (line~\ref{line:q-value}), without
actually applying the mutation. A target reward value \rtarget is used to train
the primary NN, i.e., finding the weights $\theta$ that minimise the loss
function $(\rtarget - Q(s_j, a_j; \theta))^2$ (line~\ref{line:train}).  
%
If the run terminates, i.e., if the budget assigned to the problem is finished, 
\rtarget is the same as the reward $r_t$.
Otherwise, \rtarget is estimated (line~\ref{line:q-value}) as a linear combination of the current reward $r_t$ and the predicted future reward $\gamma \hat{Q}(s_{t+1}, \hat{a}_{t+1})$, where $\hat{Q}$ is the (predicted) target Q-value and $\gamma$ is the discount factor that makes the training focus more on immediate results compared to future rewards.

Finally, the primary and target NNs are synchronised periodically by copying
the weights $\theta$ from the primary NN  to the $\hat{\theta}$ of the target NN
 every fixed number of $C$ training steps
(line~\ref{line:sync}). That is, the target NN uses an older set of weights to
compute the target Q-value, which keeps the target value \rtarget from changing
too quickly. At every step of training (line~\ref{line:train}), the Q-values
generated by the primary NN shift. If we are using a constantly shifting set of
values to calculate \rtarget (line~\ref{line:q-value}) and adjust the NN weights
(line~\ref{line:train}), then the target value estimations can easily become
unstable by falling into feedback loops between \rtarget and the
(target) Q-values used to calculate \rtarget. In order to mitigate that risk, the
target NN is used to generate target Q-values ($\hat{Q}$) that are used to compute \rtarget,
which is used in the loss function for training the primary NN. While the
primary NN is trained, the weights of the target NN are fixed.

\subsection{Online phase}

Once the learning is finished, the weights of the primary NN are frozen. In the
testing phase, the mutation strategy is selected online during an optimization
run on an unseen function. The online AOS with DE is shown in
Algorithm~\ref{algo:ddqn-test}.
Since the weights of the NN are not updated in this phase, we do not maintain a
memory of observations or compute rewards. As a new state is observed $s_t$,
the Q-values per mutation strategy are calculated and a new mutation strategy
is chosen according to the greedy policy
(line~\ref{line:test-aos}).

\begin{algorithm}[!t]
\begin{algorithmic}[1]
\STATE Initialise parameter values of DE ($F$, $NP$, $CR$) 
\STATE Initialise and evaluate fitness of each  $\xii$ in the population
\STATE Initialise $Q(\cdot)$ for each mutation strategy with fixed weights $\theta$
\STATE $t=0$ 
    \WHILE{$t < \FEmax$}
    \FOR{$i=1,\dotsc, NP$}
    \STATE Select $a_t = \argmax_a Q(s_t, a; \theta)$\label{line:test-aos}
        \STATE Generate trial vector $\ui$ for parent $\xii$ using operator $a_t$ 
        \STATE Evaluate trial vector $\ui$
        \STATE Replace $\xii$ with the best among parent and trial vector 
        \STATE $t=t+1$
        \ENDFOR
        \ENDWHILE
\RETURN best solution found
\end{algorithmic}
\caption{DE-DDQN testing algorithm}\label{algo:ddqn-test}
\end{algorithm}



\section{State features and reward}\label{proposed}
In this section we describe the new state features and reward definitions
explored for the proposed DE-DDQN method.

\subsection{State representation}

The state representation needs to provide sufficient information so that the NN
can decide which action is more suitable at the current step.  We propose a
state vector consisting of various features capturing properties of the
landscape and the history of operator performance. Each feature is
normalised to the range $[0,1]$ by design in order to abstract absolute values specific to
particular problems and help generalisation. Features are summarised in Table~\ref{tab:State-features}.

\begin{table*}[t]
  \centering
  \caption{State features}
  \label{tab:State-features}
  \smaller%
\begin{tabular*}{\textwidth}{ccm{0.56\textwidth}}
\toprule
\bf Index & \bf Feature & \bf Notes \\ \midrule
1         & $\dfrac{f(\xii) - \fbsf}{\fwsf - \fbsf}$ & $\xii$ denotes the $i$-th solution of the population and $f(\xii)$ denotes its fitness;
$\fbsf$ and $\fwsf$ denote the best-so-far and worst-so-far fitness values found up to this step within a single run
\\\midrule
2         & $\dfrac{\sum_{j = 1}^{NP} \frac{f(\xj)}{NP} - \fbsf}{\fwsf - \fbsf}$ & $NP$ is the population size \\ \midrule
3         & $\dfrac{\stdev_{j=1,\dotsc, NP}(f(\xj))}{\maxstdev}$
& $\stdev(\cdot)$ calculates the standard deviation and \maxstdev\ is the value when $NP/2$ solutions have fitness $\fwsf$ and the other half have fitness $\fbsf$\\\midrule
4         & $\dfrac{\FEmax - t}{\FEmax}$ & $\FEmax$ is the maximum number of function evaluations per run, and $\FEmax - t$ gives the remaining number of evaluations at step $t$ \\\midrule
5         & $\dfrac{\fdim}{\maxfdim}$ & $\fdim$ is the dimension of the benchmark function $f$ being optimised, and $\maxfdim$ is the maximum dimension among all training functions \\\midrule 
6         & $\dfrac{\stagcount}{\FEmax}$ & \stagcount is the \emph{stagnation counter}, i.e., the number of function evaluations (steps) without improving $\fbsf$\\\midrule
7-11      & $\dfrac{\distance{\xii - \xj}}{\maxdist}$, $\forall j \in \{r_1,r_2,r_3,r_4,r_5\}$ &$\distance{\cdot}$ is the Euclidean distance between two solutions; \maxdist is the maximum distance possible, calculated between the lower and upper bounds of the decision space; $\{r_1,r_2,r_3,r_4,r_5\}$ are  random indexes\\\midrule                                                                                                                          
12        & $\dfrac{\distance{\xii - \xbest}}{\maxdist}$ & \xbest  is the best parent in the current population\\\midrule                                             
13-17     & $\dfrac{f(\xii) - f(\xj)}{\fwsf - \fbsf}$, $\forall j \in \{r_1,r_2,r_3,r_4,r_5\}$ & \\ \midrule
18        & $\dfrac{f(\xii) - f(\xbest)}{\fwsf - \fbsf}$ & \\ \midrule
19        & $\dfrac{\distance{\xii - \xbsf}}{\maxdist}$ & $\xbsf$ denotes the solution with fitness $\fbsf$ \\\midrule
20-35     & $\sum_{g = 1}^{\maxgen}\dfrac{\Nsucc_m(g,\op)}{\Ntot(g,\op)}$ & \multirow{3}{=}{For each \op\ and   $m \in\{1,2,3,4\}$ and normalised over all operators; \maxgen is the number of recent generations recorded; $\Nsucc_m(g,\op)$ and $\Ntot(g,\op)$ are successful and total applications of \op\ according to $\OM_m$ at generation $g$ } \\\cmidrule{1-2}
%
36-51     & $\dfrac{\sum_{g=1}^{\maxgen} \sum_{k=1}^{\Nsucc_m(g,\op)} \OM_m(g, k, \op)}{\sum_{g=1}^{\maxgen} \Ntot(g,\op)}$ \\ \midrule
52-67     & $\dfrac{\BestOM_m(\maxgen,\op) - \BestOM_m(\maxgen - 1,\op)}{\BestOM_m(\maxgen - 1,\op) \cdot \lvert\Ntot(\maxgen,\op) - \Ntot(\maxgen-1,\op)\rvert}$&
For each \op\ and   $m \in\{1,2,3,4\}$ and normalised over all operators; $\BestOM_m(g,\op)$ is the maximum value of $\OM_m(g, k, \op)$\\\midrule

68-83     & $\sum_{g=1}^{\maxgen} \BestOM_m(g,\op)$ & For each \op\ and   $m \in\{1,2,3,4\}$ and normalised over all operators \\ \midrule
84-99     & $\sum_{w=1}^{W} \OM_m(w, \op)$ & For each \op\ and   $m \in\{1,2,3,4\}$ and normalised over all operators; $\OM_m(w, \op)$ is the $w$-th value in the window generated by \op  \\ \bottomrule
  \end{tabular*}
\end{table*}

Our state needs to encode information about how the current solutions in the
population are distributed in the decision space and their differences in
fitness values.
The fitness of current parent $f(\xii)$ is given to the NN as a first state feature. 
The next feature is the mean of the fitness of the current population.
The first two features in the state are normalised by the difference of worst and best seen so far solution.
%
%
The third feature calculates the standard deviation of the population fitness values.
%
%
Feature 4 measures the remaining budget of function evaluations.
%
Feature 5 is the dimension of the function being solved. The training set includes benchmark functions with different dimensions in the hope that the NN are able to generalise to functions of any dimension within the training range.
Feature 6, stagnation count, calculates the number of function evaluations since the last improvement of the best fitness found for this run (normalised by \FEmax).

The next set of feature values describe the relation between the current parent and the six solutions used by the various mutation strategies, i.e., the five random indexes ($r_1$, $r_2$, $r_3$, $r_4$, $r_5$) and the best
parent in the population ($\xbest$). Features 7--12 measure the Euclidean distance
in decision space between the current parent $\xii$ and the six solutions.
These six euclidean distances help the NN learn to select the strategy that best combines these solutions.
Features 13--18 use the same six solutions to calculate the fitness difference w.r.t. $f(\xii)$.
Feature 19 measures the normalised Euclidean distance in decision space between
$\xii$ and the best solution seen so far. We use distances instead of
positions to make the state representation independent of the dimensionality of
the solution space.



Describing the current population is not sufficient to select the best
strategy. Reinforcement learning requires the state to be Markov, i.e., to
include all necessary information for selecting an action. To this end, we
enhance the state with features about the run time history. Using historical
information has shown to be useful in our previous
work~\cite{ShaLopKaz2018ppsn}.  In addition to the remaining budget and the
stagnation counter described above,
 we also store four metric values $\OM_m(g,k,\op)$ after the application of $\op$ at generation $g$:
\begin{enumerate}
\item $\OM_1(g,k,\op) =  f(\xii) - f(\ui)$, that is, the $k$-th fitness improvement of offspring $\ui$ over parent $\xii$;
\item $\OM_2(g,k,\op)$,  the $k$-th  fitness improvement of offspring over \xbest, the best parent in the current population;
\item $\OM_3(g,k,\op)$, the $k$-th  fitness improvement of offspring over \xbsf, the best so far solution; and
\item $\OM_4(g,k,\op)$, the $k$-th  fitness improvement of offspring over the median fitness of the parent population.
\end{enumerate}
For each $\OM_m$, the total number of fitness improvements (\emph{successes}) is given by
$\Nsucc_m(g,\op)$, that is, the index $k$ is always  $1 \leq k \leq \Nsucc_m(g,\op)$.
The counter $\Ntot(g,\op)$ gives the total number of applications of $\op$ at
generation $g$. We store this historical information for the last $\maxgen$
number of generations.

With the information above, we compute the sum of success rates
over the last $\maxgen$ generations, where each success rate is the number of successful applications of operator $\op$,
i.e., mutation strategy, in generation $g$ that improve metric $\OM_m$ divided
by the total number of applications of $\op$ in the same generation. For each metric $\OM_m$, the values for an operator are normalised by the sum of all values of all operators.  
%
A different success rate is calculated for each combination of $\OM_m$
($m \in\{1,2,3,4\}$) and $\op$ (four mutation strategies) resulting in features
20--35.

We also compute the sum of fitness improvements for each $\OM_m$ divided by the
total number of applications of $\op$ over the last \maxgen generations
(features 36--51).  Features 52--67 are defined in terms of best fitness
improvement of a mutation strategy $\op$ according to metric $\OM_m$ over a
given generation $g$, that is,
$\BestOM_m(g,\op) = \max_{k}^{\Nsucc_m(g,\op)} \OM_m(g, k, \op)$. In this case, we calculate the relative difference in best improvement of the last generation with respect to the previous one, divided by the difference in number of applications between the last two generations (\maxgen and $\maxgen -1$). Any zero value in the denominator is ignored. 
The sum of best improvement seen for combination of operator and metric is given as features 68--83. 
%



Features 84-99 are calculated by maintaining a fixed size window $W$ 
where each element is a tuple of the four metric values $OM_m, m\in\{1,2,3,4\}$ and $f(\ui)$ resulting from the application of a mutation strategy to $\xii$ that generates $\ui$. Initially the window is filled with $OM_m$ values as new improved offsprings are produced. Once it is full, new elements replace existing ones generated by that mutation strategy according to the First-In First-Out (FIFO) rule. If there is no element produced by that operator in the window, the element with the worst (highest) $f(\ui)$ is replaced.
Each feature is the sum of $OM_m$ values within the window for each $m$ and each operator.
The difference between features extracted from recent generations (68-83) and from the fixed-size window (84-99) is that the window captures the best solutions for each operator, and the number of solutions present per operator vary. In a sense, solutions compete to be part of the window.
Whereas when computing features from the last \maxgen generations, all successful improvements per generation are captured and there is no competition among elements. As the most recent history is the most useful, we use small values for last $\maxgen = 10$ generations and window size  $W=50$.


\subsection{Reward definitions}
While we only know the true reward of a sequence of actions after a full run of DE is completed, i.e., the best fitness found, such sparse rewards provide a very weak signal and can slow down training.
Instead, we calculate rewards after every action has been taken, i.e., a new offspring $\vec{u}_i$ is produced from parent $\xii$. 
In this paper, we explore three reward definitions, each one using different information related to fitness improvement:
\begin{align*}
  \text{R1} & = \max \{f(\xii) - f(\ui), 0 \} &
  \text{R2} & = \begin{cases}  10  & \text{if }f(\ui)<\fbsf) \\ 1 & \text{else if } f(\ui)<f(\xii)\\0 & \text{otherwise} \end{cases}\\
  \text{R3} & = \max \{\tfrac{f(\xii) - f(\ui)}{f(\ui) - f_\text{optimum}}, 0\}
\end{align*}
R1 is the fitness difference of offspring from
parent when an improvement is seen. This definition has been used commonly in
literature for parameter control~\cite{PetEve2002control, CheGaoChen2005scga,
  SakTakKaw2010}. R2 assigns a higher reward to an improvement over the best so
far solution than to an improvement over the parent. Finally, R3 is a variant
of R1 relative to the difference between the offspring fitness and the optimal
fitness, i.e., maximise the fitness difference between parent and offspring
and minimise fitness difference between offspring and optimal solution.  This
definition can only be used when the optimum values of the functions used for training are
known in advance.


\section{Experimental design}\label{sec:exp}

In our implementation of DE-DDQN, the primary and target NNs are multi-layer
perceptrons. We integrate the three reward definitions R1, R2 and R3 into
DE-DDQN and the resulting methods are denoted DE-DDQN1, DE-DDQN2 and DE-DDQN3,
respectively. For each of these methods, we trained four NNs using batch sizes 64 or 128 and 3 or 4 hidden layers,
and we picked the best combination of batch size and number of
hidden layers according to the total accumulated reward during the training
phase. 
In all cases, the most successful configuration was batch size 64
with 4 hidden layers. Results of other configurations are not shown in the
paper.

%

The rest of the parameters are not tuned but set to typical values. In the training phase, we applied $\epsilon$-greedy 
policy with $\epsilon = 10\%$ of the actions selected randomly and the rest according to the highest Q-value. In the warm-up phase during training, we set the capacity of the memory of observations larger than the warm-up size so that 90\% of the memory is filled up with observations from random actions and the rest with actions selected by the NN. %
The gradient descent algorithm used to update the weights of the NN during training is Adam~\cite{KinBa2014adam}. 
%
Table~\ref{tab:Parameter-values} shows all hyperparameter values. 
\begin{table}[!tb]
       \centering
       \caption{Hyperparameter values of DE-DDQN\label{tab:Parameter-values}}
       \resizebox{0.9\columnwidth}{!}{%
       \begin{tabular}{rl}
             \toprule
             \bf Training and online  parameters & \bf Parameter value \\ \cmidrule(lr){1-1}\cmidrule(lr){2-2}
             Scaling factor ($F$) & $0.5$ \\
             Crossover rate ($CR$) & $1.0$ \\
             Population size ($NP$) & $100$ \\
             $\FEmax$ per function & $10^4$ function evaluations \\
             Max. generations ($\maxgen$) & $10$ \\
             Window size ($W$) & $50$ \\
             Type of neural network & Multi layer perceptron \\ 
             Hidden layers &  $4$ \\ 
             Hidden nodes & $100$ per hidden layer \\ 
             Activation function & Rectified linear (Relu)~\cite{NaiHin2010rectified} \\
             Batch size & $64$ \\ \midrule
             \bf Training only parameters & \bf Parameter value \\ \cmidrule(lr){1-1}\cmidrule(lr){2-2}
             Training policy & $\epsilon$-greedy ( $\epsilon = 0.1$) \\ 
             Discount factor ($\gamma$) & $0.99$ \\ 
             Target network synchronised ($C$) & every $1e3$ steps \\ 
             Observation memory capacity & $10^5$ \\ 
             Warm-up size & $10^4$ \\ 
             NN training algorithm & Adam (learning rate: $10^{-4}$)\\\midrule
             \bf Online phase parameters & \bf Parameter value \\ \cmidrule(lr){1-1}\cmidrule(lr){2-2}
             Online policy & Greedy \\ 
            \bottomrule
       \end{tabular}}
\end{table}

We compared the three proposed DE-DDQN variants with ten baselines: random
selection of mutation strategies (Random), four different fixed-strategy DEs
(DE1-DE4), PM-AdapSS (AdapSS)~\cite{FiaSchoSeb2010toward}, F-AUC
(FAUC)~\cite{GonFiaCai2010adaptive}, RecPM-AOS (RecPM)~\cite{ShaLopKaz2018ppsn}
and the two winners of CEC2005 competition, which are both variants of CMAES:
LR-CMAES (LR)~\cite{AugHan2005lrcmaes} and IPOP-CMAES
(IPOP)~\cite{AugHan2005cec}. Among all these alternatives, AdapSS, FAUC, RecPM
are AOS methods that were proposed to adaptively select mutation
strategies. 
The parameters of these AOS
methods 
were previously tuned with the help of an offline configurator~\textsc{irace}~\cite{ShaLopKaz2018ppsn} and the tuned hyperparameter values (parameters of
AOS and not DE) have been used in the experiments. The first eight baselines
involve the DE algorithm with the following parameter values: population size ($\textit{NP} = 100$), scaling factor ($F = 0.5$) and crossover rate
($CR = 1.0$). This choice for parameter $F$ has shown good
results~\cite{Fialho2010PhD}. 
CR as $1.0$ has been chosen to see the full potential of mutation strategies to
evolve each dimension of each
parent.
The results of LR and IPOP are taken from their original papers from the \textsc{cec}2005 competition 
for the comparison.

\begin{table*}[!ht]
    \centering%
    \caption{Mean (and standard deviation in parenthesis) of function error values obtained by 25 runs for each function on test set. Former five are dimension 10 and last five are dimension 30. We refer DE-DDQN as DDQN. Bold entry is the minimum mean error found by any method for each function.}
    \label{tab:results}
    \resizebox{\textwidth}{!}{%
    \begin{tabular}{|*{14}{c|}}
    \toprule
         Function & Random & DE1 & DE2 & DE3 & DE4 & AdapSS & FAUC & RecPM & LR & IPOP & DDQN1 & DDQN2 & DDQN3  \\ \midrule
         $F3$-10 & \begin{tabular}{@{}c@{}}2.34e+8\\(1.06e+8) \end{tabular} & \begin{tabular}{@{}c@{}} 2.78e+8\\(1.30e+8) \end{tabular} & \begin{tabular}{@{}c@{}} 2.26e+8\\(1.10e+8) \end{tabular} & \begin{tabular}{@{}c@{}} 2.38e+8\\(1.23e+8) \end{tabular} & \begin{tabular}{@{}c@{}} 2.63e+8\\(1.42e+8) \end{tabular} & \begin{tabular}{@{}c@{}}3.37e+4\\(3.62e+5)\end{tabular} & \begin{tabular}{@{}c@{}}3.53e+5\\(1.65e+4)\end{tabular} & \begin{tabular}{@{}c@{}} 3.08e+4\\(2.64e+4) \end{tabular} & \begin{tabular}{@{}c@{}} \textbf{4.94e-9}\\(1.45e-9) \end{tabular} & \begin{tabular}{@{}c@{}} 5.60e-9\\(1.93e-9) \end{tabular} & \begin{tabular}{@{}c@{}} 3.98e+3\\(1.91e+3) \end{tabular} & \begin{tabular}{@{}c@{}} 7.38 e+0\\(3.59e0) \end{tabular} & \begin{tabular}{@{}c@{}} 2.12e+1\\(1.14e+1) \end{tabular} \\ \hline
         
         $F9$-10 & \begin{tabular}{@{}c@{}} 1.20e+2\\(1.32e+1) \end{tabular}& \begin{tabular}{@{}c@{}} 1.18e+2\\(1.20e+1) \end{tabular}& \begin{tabular}{@{}c@{}} 1.22e+2\\(1.88e+1) \end{tabular}& \begin{tabular}{@{}c@{}} 1.16e+2\\(1.44e+1) \end{tabular}& \begin{tabular}{@{}c@{}} 1.22e+2\\(1.71e+1) \end{tabular}& \begin{tabular}{@{}c@{}}4.10e+1\\(6.36e+0)\end{tabular} & \begin{tabular}{@{}c@{}}4.36e+1\\(5.99e+0)\end{tabular} & \begin{tabular}{@{}c@{}} 3.79e+1\\(6.33e+0) \end{tabular}& \begin{tabular}{@{}c@{}} 8.60e+1\\(3.84e+1) \end{tabular}& \begin{tabular}{@{}c@{}} \textbf{6.21e+0}\\(2.10e+0) \end{tabular}& \begin{tabular}{@{}c@{}} 4.19e+1\\(6.21e+0) \end{tabular}& \begin{tabular}{@{}c@{}} 3.68e+1\\(4.64e+0) \end{tabular}& \begin{tabular}{@{}c@{}} 3.86e+1\\(7.66e+0) \end{tabular} \\ \hline
         
         $F16$-10 & \begin{tabular}{@{}c@{}} 6.46e+2\\(1.02e+2) \end{tabular}& \begin{tabular}{@{}c@{}} 6.50e+2\\(9.65e+1) \end{tabular}& \begin{tabular}{@{}c@{}} 6.31e+2\\(1.15e+2) \end{tabular}& \begin{tabular}{@{}c@{}} 5.91e+2\\(1.07e+2) \end{tabular}& \begin{tabular}{@{}c@{}} 6.33e+2\\(9.97e+1) \end{tabular}& \begin{tabular}{@{}c@{}}1.90e+2\\(2.21e+1)\end{tabular} & \begin{tabular}{@{}c@{}}2.05e+2\\(1.41e+1)\end{tabular} & \begin{tabular}{@{}c@{}} 1.89e+2\\(1.25e+1) \end{tabular}& \begin{tabular}{@{}c@{}} 1.49e+2\\(8.01e+1) \end{tabular}& \begin{tabular}{@{}c@{}} \textbf{1.11e+2}\\(1.66e+1) \end{tabular}& \begin{tabular}{@{}c@{}} 1.93e+2\\(1.24e+1) \end{tabular}& \begin{tabular}{@{}c@{}} 1.79e+2\\(2.05e+1) \end{tabular}& \begin{tabular}{@{}c@{}} 1.88e+2\\(1.41e+1) \end{tabular}\\ \hline
         
         $F18$-10 & \begin{tabular}{@{}c@{}} 1.33e+3\\(1.16e+2) \end{tabular}& \begin{tabular}{@{}c@{}} 1.36e+3\\(8.81e+1) \end{tabular}& \begin{tabular}{@{}c@{}} 1.39e+3\\(1.11e+2) \end{tabular}& \begin{tabular}{@{}c@{}} 1.36e+3\\(1.09e+2) \end{tabular}& \begin{tabular}{@{}c@{}} 1.36e+3\\(9.67e+1) \end{tabular}& \begin{tabular}{@{}c@{}}6.13e+2\\(1.67e+2)\end{tabular} & \begin{tabular}{@{}c@{}}6.94e+2\\(1.93e+2)\end{tabular} & \begin{tabular}{@{}c@{}} 6.48e+2\\(1.82e+2) \end{tabular}& \begin{tabular}{@{}c@{}} 8.40e+2\\(2.17e+2) \end{tabular}& \begin{tabular}{@{}c@{}}6.02e+2\\(2.76e+2) \end{tabular}& \begin{tabular}{@{}c@{}} \textbf{5.20e+2}\\(1.93e+2) \end{tabular}& \begin{tabular}{@{}c@{}} 5.81e+2\\(2.47e+2) \end{tabular}& \begin{tabular}{@{}c@{}} 5.98e+2\\(2.61e+2) \end{tabular}\\ \hline
         
         $F23$-10 & \begin{tabular}{@{}c@{}} 1.49e+3\\(5.16e+1) \end{tabular}& \begin{tabular}{@{}c@{}} 1.51e+3\\(6.71e+1) \end{tabular}& \begin{tabular}{@{}c@{}} 1.51e+3\\(6.03e+1) \end{tabular}& \begin{tabular}{@{}c@{}} 1.51e+3\\(5.58e+1) \end{tabular}& \begin{tabular}{@{}c@{}} 1.49e+3\\(4.97e+1) \end{tabular}& \begin{tabular}{@{}c@{}}6.66e+2\\(1.99e+2)\end{tabular} & \begin{tabular}{@{}c@{}}7.73e+2\\(2.05e+2)\end{tabular} & \begin{tabular}{@{}c@{}} 6.37e+2\\(1.23e+2) \end{tabular}& \begin{tabular}{@{}c@{}} 1.22e+3\\(5.16e+2) \end{tabular}& \begin{tabular}{@{}c@{}} 9.49e+2\\(3.52e+2) \end{tabular}& \begin{tabular}{@{}c@{}} \textbf{6.18e+2}\\(1.40e+2) \end{tabular}& \begin{tabular}{@{}c@{}} 6.56e+2\\(1.57e+2) \end{tabular}& \begin{tabular}{@{}c@{}} 6.90e+2\\(1.35e+2) \end{tabular}\\ \hline
         
         $F3$-30 & \begin{tabular}{@{}c@{}} 2.48e+9\\(6.60e+8) \end{tabular}& \begin{tabular}{@{}c@{}} 2.68e+9\\(7.84e+8) \end{tabular}& \begin{tabular}{@{}c@{}} 2.50e+9\\(9.04e+8) \end{tabular}& \begin{tabular}{@{}c@{}} 2.65e+9\\(6.69e+8) \end{tabular}& \begin{tabular}{@{}c@{}} 2.51e+9\\(8.22e+8) \end{tabular}& \begin{tabular}{@{}c@{}}1.52e+7\\(5.50e+7)\end{tabular} & \begin{tabular}{@{}c@{}}6.44e+7\\(5.88e+6)\end{tabular} & \begin{tabular}{@{}c@{}} 1.31e+7\\(6.84e+6) \end{tabular}& \begin{tabular}{@{}c@{}} \textbf{1.28e+6}\\(7.13e+5) \end{tabular}& \begin{tabular}{@{}c@{}} 6.11e+6\\(3.79e+6) \end{tabular}& \begin{tabular}{@{}c@{}} 1.52e+7\\(9.07e+6) \end{tabular}& \begin{tabular}{@{}c@{}} 3.06e+6\\(2.54e+6) \end{tabular}& \begin{tabular}{@{}c@{}} 5.72e+6 \\(1.30e+7) \end{tabular}\\ \hline
         
         $F9$-30 & \begin{tabular}{@{}c@{}} 5.33e+2\\(3.09e+1) \end{tabular}& \begin{tabular}{@{}c@{}} 5.27e+2\\(3.40e+1) \end{tabular}& \begin{tabular}{@{}c@{}} 5.42e+2\\(3.73e+1) \end{tabular}& \begin{tabular}{@{}c@{}} 5.19e+2\\(4.53e+1) \end{tabular}& \begin{tabular}{@{}c@{}} 5.41e+2\\(3.43e+1) \end{tabular}& \begin{tabular}{@{}c@{}}2.54e+2\\(2.69e+1)\end{tabular} & \begin{tabular}{@{}c@{}}2.88e+2\\(1.72e+1)\end{tabular} & \begin{tabular}{@{}c@{}} 2.53e+2\\(1.26e+1) \end{tabular}& \begin{tabular}{@{}c@{}} 4.19e+2\\(1.02e+2) \end{tabular}& \begin{tabular}{@{}c@{}} \textbf{4.78e+1}\\(1.15e+1) \end{tabular}& \begin{tabular}{@{}c@{}} 2.73e+2\\(1.97e+1) \end{tabular}& \begin{tabular}{@{}c@{}} 2.39e+2\\(1.52e+1) \end{tabular}& \begin{tabular}{@{}c@{}} 2.73e+2\\(2.24e+1) \end{tabular}\\ \hline
         
         $F16$-30 & \begin{tabular}{@{}c@{}} 1.19e+3\\(1.36e+2) \end{tabular}& \begin{tabular}{@{}c@{}} 1.18e+3\\(1.72e+2) \end{tabular}& \begin{tabular}{@{}c@{}} 1.18e+3\\(1.16e+2) \end{tabular}& \begin{tabular}{@{}c@{}} 1.21e+3\\(1.35e+2) \end{tabular}& \begin{tabular}{@{}c@{}} 1.20e+3\\(1.63e+2) \end{tabular}& \begin{tabular}{@{}c@{}}3.11e+2\\(6.26e+1)\end{tabular} & \begin{tabular}{@{}c@{}}3.48e+2\\(5.27e+1)\end{tabular} & \begin{tabular}{@{}c@{}} 2.97e+2\\(3.00e+1) \end{tabular}& \begin{tabular}{@{}c@{}} 2.52e+2\\(2.08e+2) \end{tabular}& \begin{tabular}{@{}c@{}} \textbf{1.96e+2}\\(1.45e+2) \end{tabular}& \begin{tabular}{@{}c@{}} 3.18e+2\\(4.22e+1) \end{tabular}& \begin{tabular}{@{}c@{}} 3.74e+2\\(9.03e+1) \end{tabular}& \begin{tabular}{@{}c@{}} 3.39e+2\\(8.41e+1) \end{tabular}\\ \hline
         
         $F18$-30 & \begin{tabular}{@{}c@{}} 1.41e+3\\(5.70e+1) \end{tabular}& \begin{tabular}{@{}c@{}} 1.43e+3\\(4.70e+1) \end{tabular}& \begin{tabular}{@{}c@{}} 1.41e+3\\(6.47e+1) \end{tabular}& \begin{tabular}{@{}c@{}} 1.42e+3\\(4.59e+1) \end{tabular}& \begin{tabular}{@{}c@{}} 1.42e+3\\(5.54e+1) \end{tabular}& \begin{tabular}{@{}c@{}}9.65e+2\\(5.59e+1)\end{tabular} & \begin{tabular}{@{}c@{}}1.02e+3\\(2.37e+1)\end{tabular} & \begin{tabular}{@{}c@{}} 9.71e+2\\(2.31e+1) \end{tabular}& \begin{tabular}{@{}c@{}} 9.64e+2\\(1.46e+2) \end{tabular}& \begin{tabular}{@{}c@{}} \textbf{9.08e+2}\\(2.76e+0) \end{tabular}& \begin{tabular}{@{}c@{}} 1.04e+3\\(2.27e+1) \end{tabular}& \begin{tabular}{@{}c@{}} 9.45e+2\\(1.42e+1) \end{tabular}& \begin{tabular}{@{}c@{}} 9.48e+2\\(3.25e+1) \end{tabular}\\ \hline
         
         $F23$-30 & \begin{tabular}{@{}c@{}} 1.58e+3\\(4.64e+1) \end{tabular}& \begin{tabular}{@{}c@{}} 1.57e+3\\(4.05e+1) \end{tabular}& \begin{tabular}{@{}c@{}} 1.55e+3\\(4.51e+1) \end{tabular}& \begin{tabular}{@{}c@{}} 1.57e+3\\(4.14e+1) \end{tabular}& \begin{tabular}{@{}c@{}} 1.57e+3\\(5.15e+1) \end{tabular}& \begin{tabular}{@{}c@{}}9.43e+2\\(1.40e+2)\end{tabular} & \begin{tabular}{@{}c@{}}1.10e+3\\(1.01e+2)\end{tabular} & \begin{tabular}{@{}c@{}} 9.67e+2\\(1.30e+2) \end{tabular}& \begin{tabular}{@{}c@{}} 7.51e+2\\(3.30e+2) \end{tabular}& \begin{tabular}{@{}c@{}} \textbf{6.92e+2}\\(2.38e+2) \end{tabular}& \begin{tabular}{@{}c@{}} 1.17e+3\\(6.30e+1) \end{tabular}& \begin{tabular}{@{}c@{}} 9.74e+2\\(1.69e+2) \end{tabular}& \begin{tabular}{@{}c@{}} 9.64e+2\\(1.70e+2) \end{tabular}\\ 
    \bottomrule
    \end{tabular}}
\end{table*}

\begin{table*}[t]
    \centering
 \caption{Average ranking of all methods.}
    \label{tab:ranking}
    \begin{tabular}{|*{14}{c|}}
    \toprule
        \bf Algo & IPOP & DDQN2 & DDQN3 & RecPM & LR & AdapSS & DDQN1 & FAUC & Random & DE3 & DE2 & DE4 & DE1 \\ \hline
        \bf Rank & 2.3 & 3.3 & 4.1 & 4.4 & 4.4 & 4.9 & 5.4 & 7.2 & 10.5 & 10.8 & 10.8 & 11.4 & 11.5\\
    \bottomrule
    \end{tabular}
   \end{table*}

\subsection{Training and testing}
In order to force the NN to learn a general policy, we 
train on different classes of functions.  From the 25 functions of the
\textsc{cec}2005 benchmark suite~\cite{SugHanLia2005cec}, we excluded
non-deterministic functions and functions without bounds (functions $F4$, $F7$,
$F17$ and $F25$). The remaining 21 functions can be divided into four classes:
 unimodal functions $F1$ -- $F5$;  basic multimodal functions $F6$ -- $F12$;
 expanded multimodal functions $F13$ -- $F14$; and
 hybrid composition functions $F15$ -- $F24$.
%
We split these 21 functions into roughly $75\%$ training and $25\%$ testing
sets, that is, $16$ functions ($F1$, $F2$, $F5$, $F6$, $F8$, $F10$--$F15$,
$F19$--$F22$ and $F24$) are assigned to the training set and the rest ($F3$,
$F9$, $F16$, $F18$ and $F23$) are assigned to the test set. According to the
above classification, the training set contains at least two functions from
each class and the test set contains at least one function from each class
except for expanded multimodal functions, as both functions of this class are
included in the training set. For each function, we consider both dimensions
$10$ and $30$, giving a total of $32$ problems for training and $10$
 problems for testing.


During training, we cycle through the 32 training problems multiple times and keep track of the mean reward achieved in each cycle. We overwrite the weights of the NN if the mean reward is better than what we have observed in previous cycles. We found this measure of progress was better than comparing rewards after individual runs, because different problems vary in difficulty making rewards incomparable. After each cycle, the 32 problems are shuffled before being used again. The mean reward stopped improving after 1890 cycles (60480 problems, $6048 \times 10^5$ FEs) which indicated the convergence of the learning process.

Although the computational cost of the training phase is significant compared
to a single run of DE, this cost is incurred offline, i.e., one time on known benchmark
functions before solving any unseen function, and it can be significantly reduced by
means of parallelisation and GPUs. On the other hand, we conjecture that
training on even more data from different classes of functions should allow the
application of DE-DDQN to a larger range of unknown functions.

After training, the NN weights were saved and used for the testing (online)
phase.\footnote{The weights obtained after training are available on
  Github~\cite{MudKomLopKaz2019gecco-supp} together with the source code, and
  can be used for testing on similar functions including expanded
  multimodal. The code may be adapted to train or test using other benchmark suites such as \textsc{bbob} with functions of up to dimension $50$.} %
For testing, each DE-DDQN variant was independently run 25 times on each test
problem and each run was stopped when either absolute error difference from the
optimum is smaller than $10^{-8}$ or $10^4$ function evaluations are
exhausted. Mean and standard deviation of the final error values achieved by
each of the 25 runs are reported in Table~\ref{tab:results}.

\subsection{Discussion of results}
The average rankings of each method among the 10 test problem instances are shown in Table~\ref{tab:ranking}. The differences among the 13 algorithms are significant ($p<.01$) according to the non-parametric Friedman test.  We conducted a post-hoc analysis using the best performing method (DE-DDQN2) among the newly proposed ones as the control method for pairwise comparisons with the other methods. The p-values  adjusted for multiple comparisons \cite{Li2008two} are shown in Table~\ref{tab:Post-hoc}.\@  The differences between DE-DDQN2 and the five baselines, random selection of operators and single strategy DEs (DE1-DE4), are significant while differences with other methods are not. The analysis makes clear that the proposed method learns to adaptively select the strategy at different stages of a DE run.

While differences between the three reward definitions are not statistically
significant, the rankings provide some evidence that R2 performs better than
the other two definitions. R2 being a simple definition assigning fixed reward values does not get affected by the function range, whereas R1 and R3 involving raw functions
values may mislead the NN when dealing with functions with different fitness
ranges.  
R2 assigns ten times more reward when offspring improves over the best so far
solution than when it improves over its parent. Thus, DE-DDQN2 may learn to
generate offspring that not only tend to improve over the parent but also
improve the best fitness seen so far. On the contrary, R1 considers the
improvement of offspring over parent only and is less informative than R3,
which considers improvement over parent and optimum value. The improvement can
be small or large when function values with different ranges is considered. As
a result, R1 and R3 become less informative about choosing operators that will
solve the problem within the given number of function evaluations. Although R3
scales fitness improvement with distance from the optimum which partially mitigates the
effect of different ranges among
functions, 
inconsistent ranges are still problematic. The
R2 definition encourages the generation of better offsprings than the best so far candidate and it is
invariant to differences in function ranges.
%
%
%
Comparing with other methods proposed in the literature shows that DE variants
with a suitable operator selection strategy can perform similarly to CMAES
variants which are known to be the best performing methods for this class of
problems.

\begin{figure}[tbp]
\begin{tabular}{@{}c@{}c@{}}
$F3$-10 & $F3$-30 \\
\includegraphics[width=0.50\linewidth]{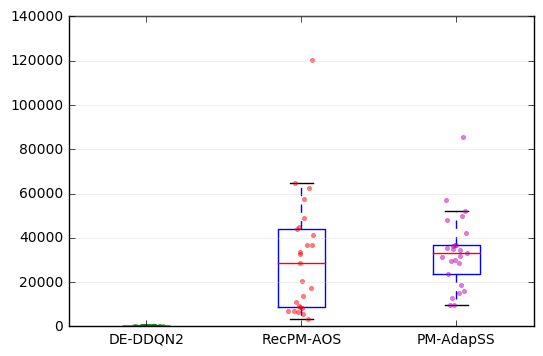} & \includegraphics[width=0.50\linewidth]{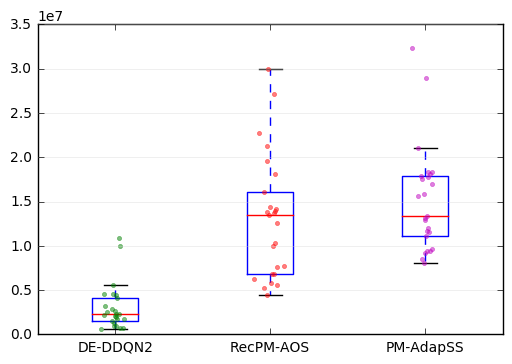} \\ 
$F9$-10 & $F9$-30 \\
\includegraphics[width=0.50\linewidth]{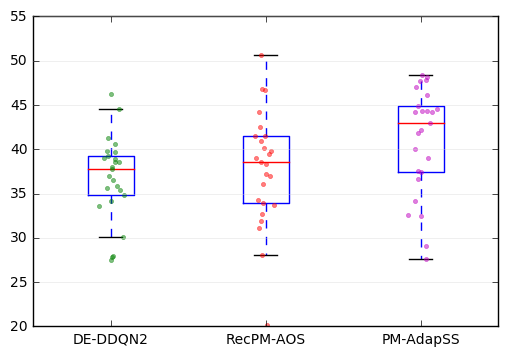} & \includegraphics[width=0.5\linewidth]{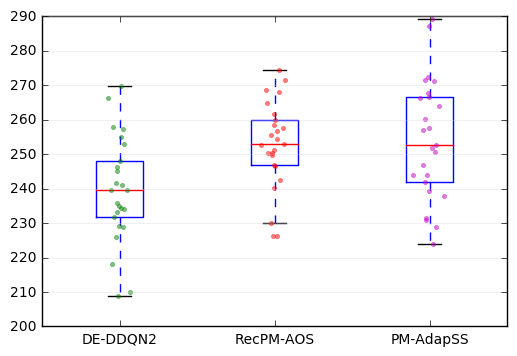} \\
$F16$-10 & $F16$-30 \\
\includegraphics[width=0.5\linewidth]{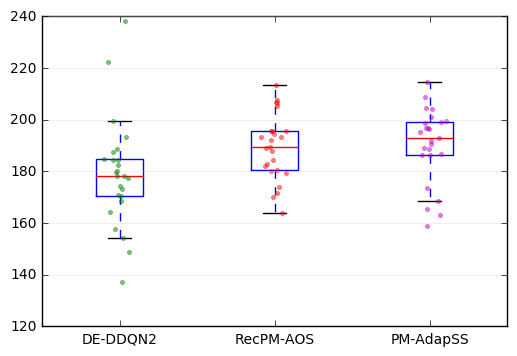} & \includegraphics[width=0.5\linewidth]{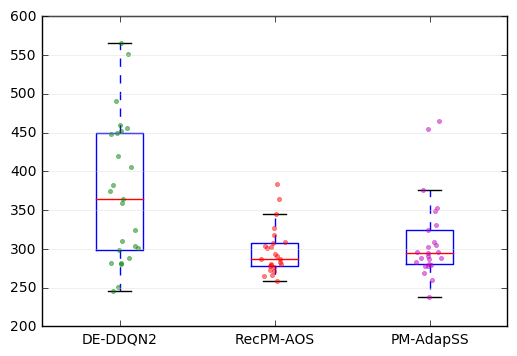} \\
$F18$-10 & $F18$-30 \\
\includegraphics[width=0.5\linewidth]{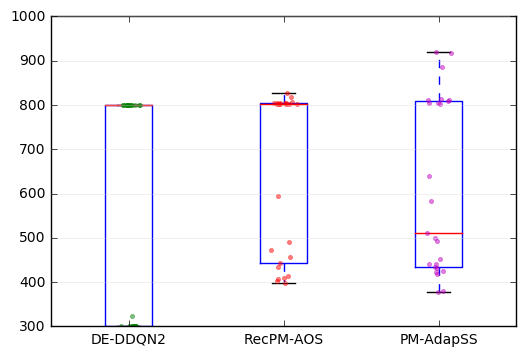} & \includegraphics[width=0.5\linewidth]{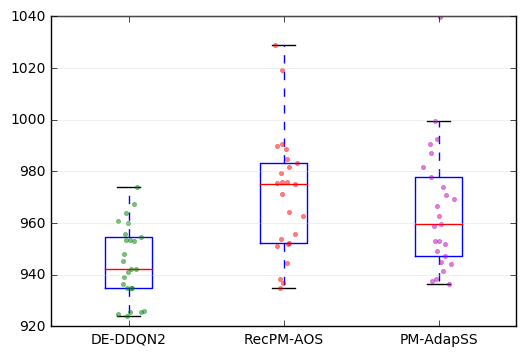} \\
$F23$-10 & $F23$-30  \\
\includegraphics[width=0.5\linewidth]{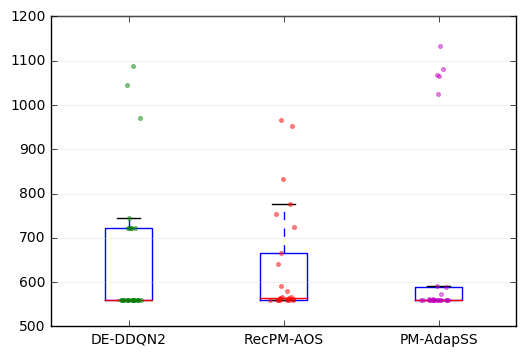} &
\includegraphics[width=0.5\linewidth]{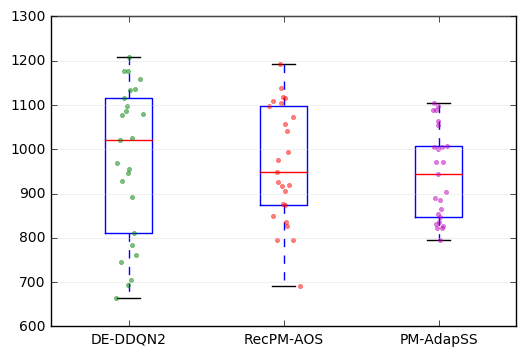} \\ 
\end{tabular}
\caption{Function error values obtained by 25 runs of DE-DDQN2, RecPM-AOS and PM-AdapSS for each function on
  test set with dimension 10 and 30.}
\label{fig:box plots}
\end{figure}

To further analyze the difference between DE-DDQN and other AOS
methods we provide boxplots of the results of 25 runs of DE-DDQN2, PM-AdapSS
and RecPM-AOS on each function (Fig.~\ref{fig:box plots}).
We observe that the overall minimum function value found across the 25 runs is lower for DE-DDQN2 in all problems except $F9$-10 and $F16$-30. As seen in box plots, for $F18$ and $F23$ with dimension 10, DE-DDQN2 often gets stuck at local optima, 
but manages to find a better overall solution compared to the other methods. Other methods find high variance solutions in these cases. 
At the same time, the median values of solutions found are better for six out of ten problems.
This observation suggests that incorporating restart strategies similar to those used by IPOP-CMAES can be particularly useful for DE-DDQN and give us a direction for future work.
DE-DDQN2 performs well consistently for the unimodal $F3$ with both 10 and 30
dimensions, while the other AOS methods find relatively higher error solutions with high variance. We interpret this as an indication that DE-DDQN can identify this type of problem and apply a more suitable AOS strategy than Rec-PM and PM-AdapSS. On the other hand, we see that for $F16$-30 and $F23$-30, DE-DDQN2 exhibits higher variance of solutions, which suggests that higher dimensional multimodal functions often confuse the NN, leading it to suboptimal behaviour.

   \begin{table}[t]
    \caption{Post-hoc (Li) using DE-DDQN2 as control method.}
    \label{tab:Post-hoc}
    \centering
    \begin{tabular}{|*{4}{c|}}
    \toprule
         \bf Comparison & \bf Statistic & \bf \begin{tabular}{@{}c@{}}Adjusted\\p-value\end{tabular} & \bf Result \\ \midrule
         DDQN2 vs DE1 & 4.70819 & 0.00001 & H0 is rejected \\ \hline
         DDQN2 vs DE4 & 4.65077 & 0.00008 & H0 is rejected \\ \hline
         DDQN2 vs DE2 & 4.30627 & 0.00005 & H0 is rejected \\ \hline
         DDQN2 vs DE3 & 4.30627 & 0.00005 & H0 is rejected \\ \hline
         DDQN2 vs Random & 4.13402 & 0.00010 & H0 is rejected \\ \hline
         DDQN2 vs FAUC & 2.23926 & 0.06630 & H0 is not rejected \\ \hline
         DDQN2 vs DDQN1 & 1.20576 & 0.39166 & H0 is not rejected \\ \hline
         DDQN2 vs AdapSS & 0.91867 & 0.50299 & H0 is not rejected \\ \hline
         DDQN2 vs Rec-PM & 0.63159 & 0.59848 & H0 is not rejected \\ \hline
         DDQN2 vs LR & 0.63159 & 0.59848 & H0 is not rejected \\ \hline
         DDQN2 vs IPOP & 0.57417 & 0.61515 & H0 is not rejected \\ \hline
         DDQN2 vs DDQN3 & 0.45934 & 0.64599 & H0 is not rejected \\ 
    \bottomrule
    \end{tabular}
\end{table}


\section{Conclusion}\label{sec:conclusion}
We presented DE-DDQN, a Deep-RL-based operator selection method that learns to
select online the mutation strategies of DE. DE-DDQN has two phases, offline
training and online evaluation phase. During training we collected data from DE
runs using a reward metric to assess the performance of the selected mutation
action and 99 features to evaluate the state of the DE. Features and reward
values are used to optimise the weights of a neural network to learn the most
rewarding mutation given the DE state. The weights learned during training are
then used during the online phase to predict the mutation strategy to use when
solving a new problem. Experiments were run using 21 functions from
\textsc{cec}2005 benchmark suite, each function was evaluated with dimensions
10 and 30. A set of 32 functions was used for training and we run the online
phase on a different test set of 10 functions.

All three proposed methods outperform all the non-AOS baselines based on mean error seen in 25 runs on test functions. This shows that the proposed methods can learn to select the right strategy at different stages of the algorithm. Our statistical analysis suggests that differences between the best proposed method and the AOS methods from the literature are not significant, but the best performing version of our model, DE-DDQN2, was ranked overall second after IPOP-CMAES. The R2 reward function, which assigns fixed reward values when better solutions are found, is more helpful for learning an AOS strategy.

For future work, we want to explore applications of Deep RL for learning to control more parameters of evolutionary algorithms, including combinations of discrete and continuous parameters. We also expect that an extensive tuning of state features and hyperparameter values will further improve performance of the method.


\newcommand{\showDOI}[1]{\unskip}   
\newcommand{\showISBN}[1]{\unskip}   
\bibliographystyle{ACM-Reference-Format-abbrv}


\providecommand{\MaxMinAntSystem}{{$\cal MAX$--$\cal MIN$} {A}nt {S}ystem}
  \providecommand{\Rpackage}[1]{#1} \providecommand{\SoftwarePackage}[1]{#1}
  \providecommand{\proglang}[1]{#1}

\end{document}